\documentclass{article}

\usepackage{arxiv}

\usepackage[utf8]{inputenc} 
\usepackage[T1]{fontenc}    
\usepackage{hyperref}       
\usepackage{url}            
\usepackage{booktabs}       
\usepackage{amsfonts}       
\usepackage{nicefrac}       
\usepackage{microtype}      
\usepackage{amsmath}
\usepackage{lipsum}         
\usepackage{graphicx}
\usepackage{natbib}
\usepackage{doi}
\usepackage{multirow}
\usepackage{algorithm}
\usepackage{comment}
\usepackage{algorithm,algpseudocode}
\usepackage{amsmath}
\DeclareMathOperator*{\argmax}{arg\,max}

\usepackage{cleveref}
\algdef{SE}[SUBALG]{Indent}{EndIndent}{}{\algorithmicend\ }%
\algtext*{Indent}
\algtext*{EndIndent}

\title{Learning to Learn with Indispensable Connections}

\author{ 
	Sambhavi Tiwari \\
	Department of Information Technology\\
	Indian Institute of Information Technology\\
	Allahabad \\
	\texttt{rsi2018503@iiita.ac.in} \\
    \And
	Manas Gogoi \\
	Department of Information Technology\\
	Indian Institute of Information Technology\\
	Allahabad \\
	\texttt{pcl2017001@iiita.ac.in} \\
    \And
	Shekhar Verma \\
	Department of Information Technology\\
	Indian Institute of Information Technology\\
	Allahabad \\
	\texttt{sverma@iiita.ac.in} \\
    \And
	Krishna Pratap Singh \\
	Department of Information Technology\\
	Indian Institute of Information Technology\\
	Allahabad \\
	\texttt{kpsingh@iiita.ac.in} \\
}

\renewcommand{\shorttitle}{\textit{arXiv} Template}

\hypersetup{
pdftitle={Learning to Learn with Indispensable Connections},
pdfsubject={q-bio.NC, q-bio.QM},
pdfauthor={David S.~Hippocampus, Elias D.~Striatum},
pdfkeywords={LTH, Meta Learning, Pruning, Few-shot learning},
}

\begin{document}
\maketitle

\begin{abstract}
	 Meta-learning aims to solve unseen tasks with few labelled instances.
Nevertheless, despite its effectiveness for quick learning in existing
optimization-based methods, it has several flaws. Inconsequential connections are frequently seen during meta-training, which results in
an over-parameterized neural network. Because of this, meta-testing
observes unnecessary computations and extra memory overhead. To overcome such flaws. We propose a novel meta-learning method called Meta-LTH that includes indispensible (necessary) connections. We applied the lottery ticket hypothesis technique known as magnitude pruning to generate these crucial connections that can effectively solve few-shot learning problem. We aim to perform two things: (a) to find a sub-network capable of more adaptive meta-learning and (b) to learn new low-level features of unseen tasks and recombine those features with the already learned features during the meta-test phase. Experimental results show that our proposed Met-LTH method outperformed existing first-order MAML algorithm
for three different classification datasets. Our method improves the classification accuracy by approximately 2\% (20-way 1-shot task setting) for omniglot dataset.

\end{abstract}


\section{Introduction}
Humans can acquire concepts through a limited number of instances, whereas machine learning models require much-annotated data. Given the expense of data annotation, it would be ideal if the model could learn from a limited number of instances. Researchers design a new technique known as meta-learning to achieve this human-level knowledge-capturing idea. Meta-learning solves the few-shot learning problem, which is the process of learning new concepts from a small number of examples.
Various applications have used meta-learning, like computer vision, reinforcement learning, architectural search, etc. Among all the machine learning applications, the most common is multi-class image classification. Meta-learning methods such as optimization-based meta-learning optimize the model for performing classification such as MAML \cite{finn2017model}, FOMAML \cite{nichol2018first}, ANIL \cite{raghu2019rapid}, Reptile \cite{nichol2018first}, iMAML \cite{rajeswaran2019meta} and others. 
This method generalizes better than RNN-based techniques  \cite{santoro2016meta} \cite{9986399} because the meta-learner learns with a logical gradient-based learning process even when it gets lesser data.
Another method known as metric-based meta-learning learns a metric space for predictions \cite{vinyals2016matching, gogoi2022adaptive}.

Amidst all, MAML is the most famous strategy for learning how a network should be initialized before fine-tuning when adapted to new tasks. But while enabling quick adaptation, MAML does not provide a compact model since it optimizes the network parameters without changing the original model architecture \cite{finn2017model}. During meta-adaptation, it is required to meta-train an over-parameterized neural network which is not necessary. To prove this, we proposed a lightweight meta-learning algorithm which performs few-shot image classification by removing irrelevant connections during meta-training.

Network pruning is a technique used in deep neural networks to eliminate redundant connections, nodes and weights. A variant of network pruning known as magnitude pruning is a technique where individual layer weights are removed during pruning to reduce the size of the model \cite{guo2016dynamic, han2015deep, hassibi1993optimal, lecun1989optimal}. It postulates that when a network is established, there should be an ideal subnetwork that can be learnt by pruning performs same or even better than the original network. The idea is that while all parameters within a neural network work together to calculate its output, many parameters can be pruned without heavily damaging the networks' accuracy. Thus, by pruning a trained network, one can achieve a model up to 90\% smaller, with negligible effect on the network's ability to perform. Hence, it results in less computation and memory overhead. 

Therefore, our proposed work performs few-shot multi-class image classification using a simple optimization-based meta-learning technique and a network pruning algorithm. As we propose a lightweight meta-learning algorithm, we used a first-order optimization-based meta-learning algorithm(FOMAML) \cite{ravi2017optimization} and a magnitude-based network pruning algorithm known as magnitude-Lottery Ticket Hypothesis \cite{elesedy2020lottery} to perform meta-training. For meta-testing, we recombine the knowledge of new test images with already learnt features by un-freezing the pruned connections. The proposed meta-learning method has been evaluated on different datasets, such as Omniglot, MiniImagenet and FC100, with different task settings. The suggested meta-learning technique is pretty simple, and experiments showed better accuracy than the benchmark FOMAML algorithm.

\section{Related Work}

Gradient-based meta-learning algorithms and their variants \cite{nichol2018first, oh2020boil, raghu2019rapid, ravi2017optimization, finn2017model} are fit for fast adaptation. These meta-learning methods often learn a strong hypothesis that can be quickly modified to unseen tasks. Optimization-based meta-learning algorithms are more popular than non-parametric \cite{koch2015siamese, snell2017prototypical, gogoi2022adaptive, sung2018learning} and black-box \cite{santoro2016meta, mishra2017simple} meta-learning algorithms. The most widely used optimization-based meta-learning algorithm is MAML \cite{finn2017model}. MAML is a straightforward and more generalized technique compared to other meta-learning strategies. Many researchers came up with several variations in the MAML technique like CAML \cite{ur2023caml}, MAML++ \cite{antoniou2018train}, and ANIL \cite{raghu2019rapid}. But the major drawback of the MAML algorithm is the computation of the Hessian-vector product during back-propagation, which is computationally demanding. To overcome this, a few other variants were proposed, like Reptile \cite{nichol2018reptile} and FOMAML \cite{nichol2018first} , ignoring the second-order derivatives in MAML. Considering simplicity, effectiveness and adaptability, FOMAML is better than others.

 Besides fast adaptation to a new task, recent research focuses on optimizing the initial architecture\cite{elsken2020meta} by integrating MAML with neural architecture. These studies aim to achieve higher few-shot inference accuracy, which may result in a model with more parameters than the initial architecture in MAML \cite{finn2017model}. However, a lot of research has been done to compress neural networks also. Some examples of these approaches are quantization \cite{hou2018loss, zhuang2019structured, zhou2018explicit}, constructing compact networks \cite{sandler2018mobilenetv2, zhang2018shufflenet, iandola2016squeezenet}, and pruning \cite{he2017channel, han2015deep}. This article focuses on sparsifying the initial architecture in optimization-based meta-learning methods, which is done using a pruning mechanism. Liu et al. \cite{liu2019metapruning} were among the first to explore the combination of meta-learning and neural network pruning. Their work aimed to train a network that could produce a collection of channel-pruned architectures along with their associated weights. These architectures and weights had to meet certain layerwise sparsity requirements provided as input. A closely related work \cite{tian2020meta} done in the past performs meta-learning and network pruning to handle meta-overfitting of optimization-based meta-learning methods. Specifically, in \cite{tian2020meta} they have combined algorithmic framework of Reptile \cite{nichol2018reptile} with network pruning, as well as two instantiations that use the network pruning subroutines Dense-Sparse-Dense (DSD) \cite{han2016dsd} and Iterative Hard Thresholding (IHT) \cite{jin2016training}, respectively to alliviate meta-overfitting. 
 
 Our approach also combines meta-learning with pruning but with a different objective. Specifically, when used for few-shot classification tasks, our goal is to sparsify (prune) the initial architecture and reuse (recombine) the learned features without sacrificing its potential for quick adaptation. 

\section{Background}
\label{sec:headings}

\subsection{\textbf{Meta-learning Foundation}}
The idea of learning how to use the knowledge obtained from performing tasks in the past to learn a new task quicker or more efficiently is known as Meta-learning. This approach solves the meta-objective (equation 1) to find an optimal meta-parameter $\theta^{*}$ using meta-training dataset $D^{tr}$ with randomly initialized parameters $\theta$.

\begin{equation}
    \theta^{*}=\argmax _{\theta} p(\theta |  D^{tr})
\end{equation}

Meta-learning is broadly divided into three broad categories: (1)Optimization based meta-learning, (2) Metric based meta-learning \cite{koch2015siamese, sung2018learning, vinyals2016matching, snell2017prototypical, gogoi2022adaptive} and (3) Memory based meta-learning \cite{santoro2016meta, munkhdalai2017meta, tiwari2022meta}. Among all, optimization-based meta-learning algorithms
are attracting more attention due to their ease of use, adaptability, and efficiency. 
This technique learns a strong hypothesis that can be quickly adapted to unseen tasks \cite{finn2017model, raghu2019rapid, ravi2017optimization}.

\subsection{\textbf{Model agnostic meta-learning}}
An example of a gradient-based meta-learning approach that optimizes the hyperparameters that arise in the gradient descent procedure for k-shot learning is MAML \cite{finn2017model}.
MAML is a gradient-based meta-learning technique which consists of two optimization loops: First, the \textbf{inner optimization} stage, also known as the task adaptation process, where each task gets adapted using its support set. Second, the \textbf{outer optimization} stage, also known as the meta updation process, where each task gets updated using its query set. The MAML update is given below:
\begin{equation}
    \min_{\theta} \sum_{\mathcal{T}_{i} \sim p(\mathcal{T})} \mathcal{L}_{\mathcal{T}_{i}}(f_{\theta^{'}_{i}}) = \sum_{\mathcal{T}_{i} \sim p(\mathcal{T})} \mathcal{L}_{\mathcal{T}_{i}}(f_{\theta_{i} - \alpha \nabla_{\theta} \mathcal{L}_{\mathcal{T}_{i}} (f_{\theta}) })
\end{equation}\\

\begin{algorithm}[h]
   \caption{FOMAML Algorithm}
   \label{alg:algo}
\begin{algorithmic}[1]
   \State {\bfseries Input:} $m$ meta-training tasks $T_{m} = {T_{1},T_{2},...,T_{m}}$ from $p(D^{tr})$ distribution, inner loop learning rate $\alpha$, outer loop learning rate $\beta$, task batch size '$s$'.
   \State {\bfseries Output:} the updated $\theta$.
   \State Initialize the network $f$ with random $\theta$.
   \State Sample mini-batch tasks $\{T_{i}\}^{s}_{i=1}$ of size $s$ from meta training tasks $T_{m}$.
   \For{ each task  $T_{i}$ }
    \State Evaluate $\nabla_{\theta}\mathcal{L}_{D^{sup}_{T_{i}}}(f_{\theta})$
    \State Compute adapted parameters using Stochastic Gradient Descent:
    \State $\phi_{i}$ = ${\theta - \alpha \nabla_\theta \mathcal{L}_{D^{sup}_{T_{i}}}(f_{\theta})}$
   \EndFor
   \State  Update $\theta^{'} \leftarrow \theta - \beta\sum_{T_{{i}\sim p(T)}}\nabla_{\theta_{i}} \mathcal{L}_{T_{i}}(f_{\phi_{i}}) $
   
\end{algorithmic}
\end{algorithm}

\subsection{\textbf{First Order MAML}}

Second derivatives are necessary for the MAML stage of meta-optimization. First-Order MAML is a modified form of MAML where instead of backward traversing the trajectory of the task-specific parameter to find the gradient of the cumulative outer loss w.r.t initial meta-parameters, we consider the gradient direction of the last task-specific update as the optimal direction of our meta-parameters. Therefore, this omits the hessian calculation, which is more compute-intensive. As a result, we develop a meta-learning algorithm that requires less computing while maintaining its effectiveness for few-shot learning.

\subsection{Almost No Inner Loop}
Raghu et al.\cite{raghu2019rapid} suggested that feature reuse alone may achieve the same quick learning performance as MAML. Researchers developed ANIL (almost no inner loop), a MAML-simplified method that is similarly effective but computationally quicker. In this work, we worked on the concept of first-order MAML. Therefore, restricting the inner loop update during meta-testing in FOMAML leads to FOANIL.

\subsection{\textbf{Lottery Ticket Hypothesis}}

Pruning has been widely studied over decades, which reduces the size of a large deep neural network model while preserving its performance. Recently, Lottery Ticket Hypothesis(LTH) \cite{frankle2018lottery, wen2016learning, han2015learning, han2016dsd, li2016pruning, liu2018rethinking} convincingly proves that a dense neural network with random initialization contains a subnetwork that, when trained independently, can match the test accuracy of the original network for at most the same amount of iterations. The most common pruning methods are magnitude pruning which searches for the winning tickets by pruning the model weights.   Another variant is the iterative magnitude pruning \cite{bai2022dual,burkholz2021existence}, which requires a huge training cost to find winning tickets iteratively. 

In weight magnitude LTH, the winning tickets are obtained by pruning the network weights whose magnitude(absolute value) is less than or equal to the user-defined threshold(pruning percentage) $p\%$. A convolution neural network with randomly initialized parameters $f(\theta)$, where $\theta$ is the weight of the network parameters. The initialized network $f(\theta)$ is trained until convergence in the initial iteration. After that, the mask $M \in \{0,1\}$ is produced by deleting the $p\%$ weight parameters with the lowest absolute value. Deleting certain connections may create subnetworks $f(\theta \odot M)$ given a pruning mask $M$. We re-initialize the network with $\theta$ initial parameters for finetuning and retrain the model with closed connections obtained with the mask $M$.

\section{Problem definition}

Two important findings are observed in meta learning: First, all the connections in a deep neural network do not contribute to prediction. To perform learning with limited data, we need to find a sparse network which can learn similarly close to or better than the original network. It is also possible that a client that needs quick adaptation may have limited resources. Second, In the meta-testing phase, we may encounter new unseen tasks; therefore, there is a need to learn new low-level features and combine them with the existing features for that, only feature re-use and re-combination will not be sufficient for rapid meta-learning.

We then discuss the few-shot learning setup and terminology used, and in the next section, we will see how our suggested approach resolves these two issues.

\subsection{Few-shot learning setup}
Given data $D$ = \{ ${D^{tr}},{D^{ts}}$ \} chosen from a distribution $p(D)$, where ${D^{tr}}$ stands for 'meta-training dataset' and ${D^{ts}}$ stands for 'meta-
testing dataset'. Both datasets have disjoint sets of classes. We consider the meta-learning problem setup as an N-way k-shot task problem. We sample $N$ classes from dataset $D$, and then $k$ samples are sampled from each class. To learn the prior and subsequently adapt to the new class instance, the model \textit{f} with randomly initialised parameters $\theta$ is trained on $m$  batches of tasks ${T}_m= \{T_1, T_2,..., T_m\}$ from the meta-train dataset ${D^{tr}}$. For evaluation during meta-testing draw $n$ batch of unseen tasks \{$ T_1,T_2,...,T_n$ \} from meta-test dataset ${D^{ts}}$. Each task $T_{i}=\{D^{sup},D^{qr}$\} is a set of \textit{support data points} $D^{sup}$ and \textit{query data points} $D^{qr}$. However, during meta-training, we have tasks with data-label pair \{$x_{i},y_{i}$\} for both support $D^{sup}$ and query set $D^{qr}$ but during meta-testing, we have tasks with labelled support set only. The query points of the meta-test task are used to evaluate the model's performance.

\section{Proposed Method}
In this section, we will discuss our proposed Meta-LTH method. This is a simple meta-learning method that finds a sparse subnetwork to conduct meta-training and meta-adapting new-unseen tasks using the notion of feature learning and recombination. Further, we will discuss the basic setup used for few-shot classification and discuss how the proposed algorithm works efficiently without hampering the meta-learning performance.
\begin{figure*}[ht]
\centering
\includegraphics[width=13cm,height=4cm]{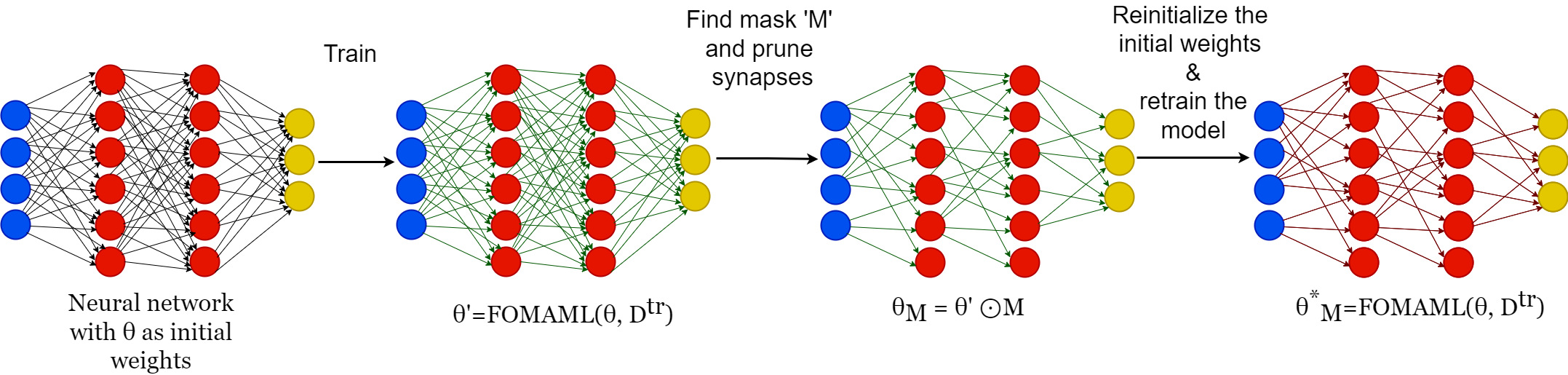}
\caption{\textbf{Schematic view of proposed Meta-LTH training phase::} From left to right, (a) The model is the randomly initialized neural network (black connections) with two hidden layers. (b) The randomly initialized neural network is trained with the meta-training dataset using the first-order MAML technique(green connections). (c) After training the neural network, find the zero-one mask $M$ w.r.t the threshold value $Th$ and prune the connections(removed green connections). (d) Finally, reinitialize the network with initial weights $\theta$ to the leftover connections and retrain the pruned network(red connections) to get the optimal meta-parameters ${\theta^{*}}_{M}$. }
\label{fig:train}
\end{figure*}

\subsection{Algorithm}
\begin{figure*}[h]
\centering
\includegraphics[width=10cm]{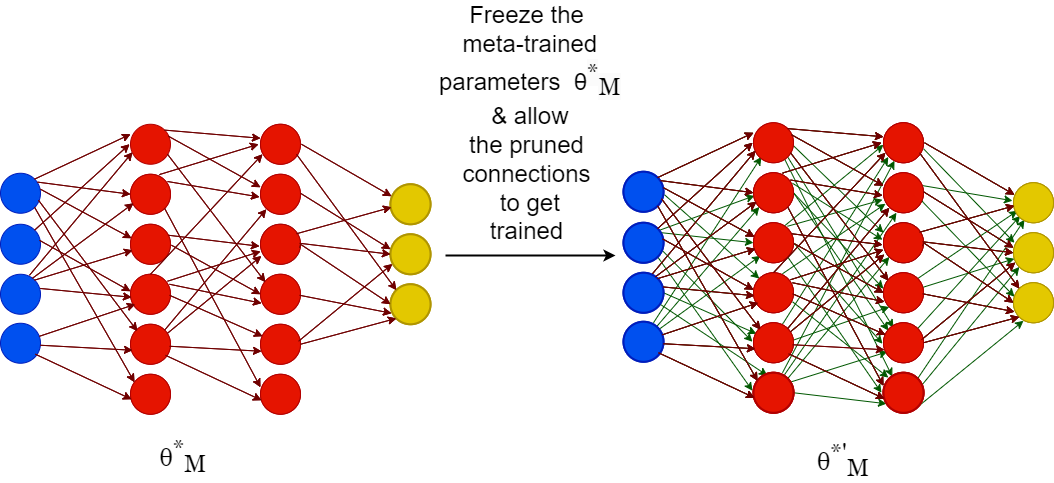}
\caption{\textbf{Schematic view of proposed Meta-LTH testing phase:} Initialize the optimal meta-parameters ${\theta^{*}}_{M}$ to the base model and allow partial fine-tuning by opening the pruned connections to get trained. The rest connections are freezed to perform feature recombination(red links are frozen, and green links are allowed to get trained). }  
\label{fig:test}
\end{figure*}

Meta-LTH is a unique gradient-based meta-learning algorithm, a simple few-shot learning technique. The motive of Meta-LTH is to find the best sparse Neural Network structure for k-shot learning using Lottery Ticket Hypothesis (LTH) within a randomly initialized Neural Network. Since Meta-LTH delivers accuracies with relatively low computing overhead, this technique competes with the benchmark gradient-based algorithm MAML \cite{finn2017model}. Instead of explicitly encoding data from a meta-training dataset into the neural network parameters as MAML does, Meta-LTH finds connections between neurons which are important for adapting to new tasks. For simplicity, in our proposed algorithm, we perform training and testing using the FOMAML algorithm.
Our work modifies the meta-training, and meta-testing phases, unlike other pruning-based meta-learning algorithms \cite{tian2020meta,liu2019metapruning}. The algorithms of our Meta-LTH are outlined in Algorithm.\ref{alg:algo(a)} for meta-training (fig.\ref{fig:train}) and Algorithm.\ref{alg:algo(b)} for meta-testing (fig.\ref{fig:test}). 
\begin{algorithm}[ht]
   \caption{Proposed Algorithm(a) Meta-Training}
    
   \label{alg:algo(a)}
\begin{algorithmic}[1]
   \State {\bfseries Input:} (a)$m$ meta-training tasks $T_{m} = {T_{1},T_{2},...,T_{m}}$ from $p(D^{tr})$ distribution, (b) Pruning percentage $p\%$.
   \State {\bfseries Output:} the updated $\theta$.
   \State Initialize a $l$ layer network $f$ with random $\theta$.
   \Indent
   \State $\theta^{'} \leftarrow FOMAML(\theta, D^{tr})$ /* Update model parameters*/
    \EndIndent
   \State Caculate threshold $Th$:   
   \Indent
         \State $Th \leftarrow percentile(\theta^{'},p)$ /* w.r.t. layer weights of the network except the classifier layer*/
         \EndIndent
    \State Generate the network zero-one mask $M$ w.r.t $\theta^{'}$ where the non-zero entries in $M$ correspond to the indices that have values greater than $Th$.
  
        \State Compute: $ \theta^{'}_{M} = \theta \odot M$ /* element-wise product */
   \State $\theta^{*}_{M} \leftarrow FOMAML(\theta_{M}, D^{tr})$, where $\theta_{M}$ is the subnetwork obtained in previous step.

\end{algorithmic}
\end{algorithm}

\textbf{Meta-Training}: During the meta-training phase, Our proposed model Meta-LTH learns the patterns of different tasks, then performs magnitude pruning on the weights and retrains the network with some closed inefficient connections. The meta-training process is explained in three steps:

\textbf{\textit{(a)Model pre-training:}} First, to pre-train the model, we run FOMAML iterations to obtain a good initialization. To execute FOMAML (Algorithm.(\ref{alg:algo})), we sample a mini-batch of task $\{T_{i}\}^{s}_{i=1}$ from meta-training dataset ${D^{tr}}$. For each task $\{T_{i}\}$ compute the gradients  of the model parameter $\nabla_{\theta}\mathcal{L}_{T_{i}}(f_{\theta})$ , where $\theta$ is the initial random weights of the model. Then compute the adapted parameters using SGD(stochastic gradient descent) for task $T_{i}$ using the following update:
\begin{equation}
    \phi_{i} = {\theta - \alpha \nabla_\theta \mathcal{L}_{T_{i}}(f_{\theta})}
\end{equation}
 where, $\phi^{'}_{i}$ is known as task-specific parameters learned from support set of each task $T_{i}$. When all the task-specific parameters are updated, then the initial parameter $\theta$ is updated according to $\theta^{'} \leftarrow \theta - \beta\sum_{T_{{i}\sim p(T)}}\nabla_{\theta_{i}} \mathcal{L}_{T_{i}}(f_{\phi_{i}}) $ with $\alpha$ and $\beta$ learning rates. 

\textbf{\textit{(b)Model Pruning:}} After the model is pre-trained, we prune the meta-trained networks in layer-wise pruning manner \cite{dong2017learning}. To find the meta-trained sub-network, Lottery Ticket Hypothesis technique known as magnitude pruning is applied to ${l-1}$ hidden layers. This method finds the "winning tickets" or the significant weights in the neural network based on their magnitude. For this, the network is trained to convergence, then pruning a subset of the weight using a certain threshold value '$Th$'. We intentionally remove all weights below a certain threshold, '$Th$', presuming that these weights do not incorporate significant feature learning. Figure \ref{fig:train} shows the schematic view of the proposed meta-training. The green connections of the network show the weight connections obtained after the pre-training phase. Afterwards, zero-one mask $M$ is computed using the threshold value $Th$ and then perform elementwise product: $\theta^{'} \odot M$ on the pre-trained weights '$\theta^{'}$'. We get a pruned network  $ \theta_{M} $, with some green links(unpruned). After obtaining the most significant weight connections(green unpruned links), we retrain the model $\theta_{M}$.

\textbf{\textit{(c)Model Retraining:}}  Finally, we retrain the model $\theta_{M}$ with FOMAML(Algorithm \ref{alg:algo}). Again we sample meta-training tasks from meta-training dataset $D^{tr}$ and follow some major constraints:\\
a) Allow only the sub-network $\theta_{M}$ to be trained.\\
b) Re-Initialize the sub-network $\theta_{M}$ with the initial $\theta$ weights.\\
After this stage, we obtain a good set of meta-parameters presuming it to be optimal for meta-adaptation.

\textbf{Meta-Testing}: After the meta-training phase, we get the meta-initialization parameters($\theta^{*}_{M}$) to perform meta-testing for unseen meta-test tasks sampled from $D^{ts}$ dataset. To understand this process of fine-tuning, see figure.\ref{fig:test}. During meta-training, we found some insignificant connections in the network, which is not updated during the training phase and shows that only some links of the neural network are responsible for learning new features. However, during the meta-adaptation phase, we encounter tasks of different classes that are never seen during the meta-training phase. Since we will be performing feature recombination with the meta-trained and new class features, we observed that the closed edges (brown edges) need to be updated during meta-retraining.\\
In contrast, the edges permitted to train during meta-retraining (blue edges) do not need to be updated further during the adaptation phase. Therefore, we proposed a novel method of fine-tuning the initialised weights to reduce the parameter overhead caused during the adaptation phase(Algorithm.\ref{alg:algo(b)}). The update equation during the meta-testing phase is shown:
\begin{equation}
    \theta^{*'}_{M} \leftarrow \theta^{*}_{M} + \nabla\theta^{*}_{M} \odot M^{comp}   
\end{equation}

Where, $\theta^{*'}_{M}$ is updated model parameter, $M^{comp}$ is the complement of mask $M$ obtained during meta-training using LTH, $\nabla\theta^{*}_{M}$ is the gradients of meta-trained parameter.

\begin{algorithm}[ht]
   \caption{Proposed Algorithm(b) Meta-Testing}
   \label{alg:algo(b)}
\begin{algorithmic}[1]
   \State {\bfseries Input:} (a)$n$ meta-training tasks $T_{n} = {T_{1},T_{2},...,T_{n}}$ from $p(D^{ts})$ distribution, (b) Meta-trained parameters $\theta^{*}_{M}$.
   \State {\bfseries Output:} Test accuracy.
   \State Initialize the network with $\theta^{*}_{M}$ parameters.
   \State Generate zero one gradient mask $M^{comp}$ such that ones entries in $M^{comp}$ corresponds to zero entries in mask $M$ and zero entries in $M^{comp}$ corresponds to ones entries in mask $M$.
   \For{ all test task $T_{i}$}
   \State Evaluate $\nabla \theta^{*}_{M} \leftarrow \nabla\underset{\theta^{*}_{M}}{L}(\theta^{*}_M,D^{sup})$
   \State Update the network parameters:
   \State $\theta^{*'}_{M} \leftarrow \theta^{*}_{M} + \nabla\theta^{*}_{M} \odot M^{comp}$ /* New feature learning through pruned connections and re-combining with already learnt features*/
   \EndFor
\State Compute Accuracy
\end{algorithmic}
\end{algorithm}

\section{Results and Discussion}
\subsection{Datasets}
We perform few-shot learning experiments on the Omniglot \cite{lake2015human} miniImageNet \cite{vinyals2016matching} and Fewshot-CIFAR100 (FC100) \cite{oreshkin2018tadam} benchmarks. MiniImageNet and Omniglot are frequently used in several few-shot learning methods[]. FC100 is a new dataset for few-shot learning setup proposed in \cite{oreshkin2018tadam} and is different from other image datasets like miniImageNet due to its lower picture resolution, and stricter training-test splits.

\subsubsection{\textbf{Omniglot}}
This dataset contains 50 different alphabets divided into 1623 handwritten characters known as classes. Each class has 20 black and white images of size 28X28 drawn by 20 persons. The images are labelled with the name of the corresponding character and a suffix. For, e.g. the Alphabet of the Sanskrit language in the dataset has 42 characters, and there are 20 images of each character with labels. In our classification task, each character is considered a separate class irrespective of language. These classes are split into training and test sets: 1200 classes for training and 423 for testing. All the character images are first augmented by performing rotations to create more data samples and reduce overfitting. Figure 1 illustrates some Hebrew language characters.
\begin{figure*}[h]

\includegraphics[width=6cm]{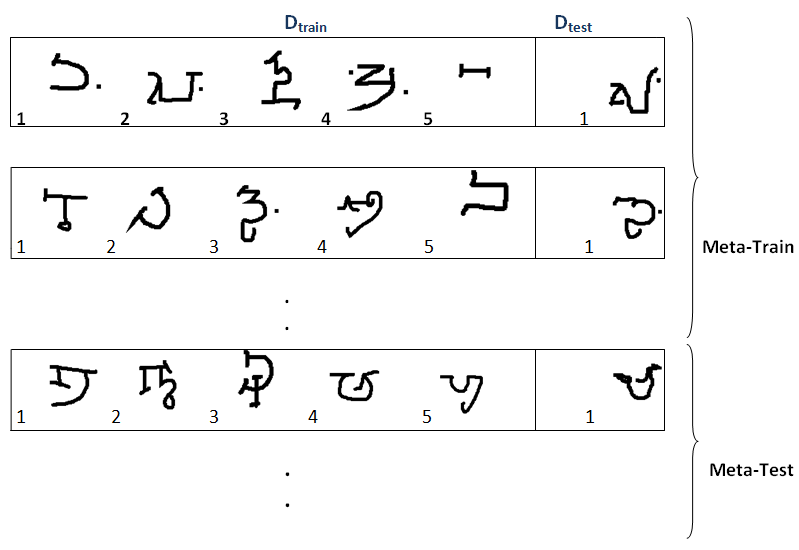}
\centering
\caption{Three sets of different character depicting three classes from Omniglot dataset. ${char1}$, $char4$ and $char16$ in Hebrew. Each character is written by $20$ different person.}

\end{figure*}

\textbf{Implementation details:} To evaluate our proposed model, we perform our experiments on the Omniglot dataset \cite{lake2015human} for classification with the help from the Torchmeta library developed by Deleu et al. [2019] \cite{deleu2019torchmeta}.

\subsubsection{\textbf{Miniimagenet}}
Ravi and Larochelle proposed the MiniImagenet dataset in 2016 \cite{vinyals2016matching}. The dataset comprises 64 training classes, 24 tests and 12 validation classes. We consider four task settings, i.e., 5-way 1-shot, 5-way 5-shot, 10-way 5-shot and 10-way 1-shot on this dataset. Therefore, we have meta-train and meta-test tasks to classify among 20 randomly chosen classes, given only a few labelled samples, i.e., 5 and 1 instance of each class. The experiments are done for 5-way 1-shot and 5-way 5-shot task settings.

\textbf{Implementation details:} All experiments were carried out on the Pytorch platform with help from the Torchmeta library developed by Deleu et al. [2019] \cite{deleu2019torchmeta} to provide a standard FOMAML implementation on this dataset.
\subsubsection{\textbf{FC100}}
It is based on the well-known object classification dataset CIFAR100 \cite{krizhevsky2009learning}. This dataset contained images of 100 different classes from the CIFAR100 dataset and was introduced first in \cite{oreshkin2018tadam}. It has lower
image resolution and more challenging meta-training/test
splits that are separated according to object superclasses.
FC100 has 100 object classes, each with 600 images of 32x32 colour resolution. These 100 classes belong to 20 superclasses from the CIFAR100 dataset. Of 100 classes, 60 classes are reserved for meta-training, 20 classes are for meta-validation, and rest 20 classes are for meta-testing data. The splits were made with the help of torchmeta \cite{deleu2019torchmeta} library. These meta-train, validation and meta-test data splits are over 12, 4 and 4 superclasses. Therefore, this split setting minimises the information overlap among each other.

\textbf{Implementation details:} We used the public code of first-order MAML \cite{deleu2019torchmeta} to get the benchmark details on this new dataset. \\

\subsection{\textbf{System Configuration}}
We have performed all our experiments on 2 GPU servers with the following configurations:\\
2x Tesla-V100 : 16GB , RAM : 32 GB, CUDA cores = 10,240, CUDA version: V9.2.148, System type : 64-bit Operating System, x64-based processor.

\subsection{Experimental Results}
In this section, we describe the experimental details and results achieved with our proposed methodology and make
comparisons with the original FOMAML \cite{nichol2018first} approach. Initially, the model is meta-trained with training tasks sampled from the meta-train dataset till convergence. The
parameters are saved to perform magnitude pruning using LTH and then re-training the sub-network with 6000 batches of training task. Finally, the trained sub-network performs feature learning and recombination using 100 batches of new unseen tasks. Three distinct datasets(Omniglot, Miniimagenet, FC100) are used to evaluate and compare the proposed methodology's performance. All experiments were carried out on the Pytorch platform with help from the Torchmeta \cite{deleu2019torchmeta} library. We ran each experiment with three different random seeds, and computed the confidence intervals using the standard deviation across the runs.

\textbf{Network Architecture:} We employed a 4CONV model \cite{vinyals2016matching} architecture in this study. 4CONV model is a popular model used in few-shot learning methods \cite{vinyals2016matching,ravi2017optimization,mishra2017simple,finn2017model}. The 4 layers, 3X3 convolutions, 32 filters, batch normalization, ReLU nonlinearity, and 2X2 max-pooling constitute the 4CONV model architecture. Each block has 64 input and 64 output hidden channels, except for the first block, which has 1 or 3 input channels depending on the number of RGB channels in the input image. The reason behind using this model architecture is its simplicity. Since our main concern for this research is to minimize the computational cost without hampering its performance on a few-shot problem setting. 

\textbf{Few-shot Classification details:}
For CIFAR100: During meta-training and retraining, the inner learning rate is 0.4, and the outer learning rate is 0.001. For meta-testing, the inner learning rate is set to 0.01, and the outer learning rate is set to 0.001. The batch size for meta-training is set to 16, and for meta-testing, it is set to 12. In both algorithms, only 10 steps of stochastic gradient descent are applied during an evaluation phase. The comparison table shows the classification accuracy (\ref{table:1}).\\
\begin{table}[ht]
\begin{center}
\begin{tabular}{| c | c | c | c | c |}
\hline
\textbf{Method} & \textbf{Prune \%} & \textbf{5way-1shot} &\textbf{5way-5shot}   \\ 
\hline
FOMAML &0 &$37.76\pm0.15$ & $50.06\pm1.03$   \\ \hline
Meta-LTH &90& \textbf{38.45$\pm $ 0.25} & \textbf{52.02 $\pm$ 0.07}\\ \hline
Meta-LTH &80& 36.94 $\pm$ 0.21 &  49.02 $\pm$ 1.09  \\ \hline
Meta-LTH &70& 36.88 $\pm$ 1.05 &  48.78 $\pm$ 0.91  \\ \hline
Meta-LTH &60& 36.46 $\pm$ 0.32 &   47.96 $\pm$ 1.11\\ \hline
Meta-LTH &50& 36.36 $\pm$ 1.61 &   47.68 $\pm$ 0.03  \\ \hline

\end{tabular}
\caption{ Classification accuracies for proposed Meta-LTH approach in comparison with FOMAML on FC100 dataset.}
\label{table:1}
\end{center}
\end{table}

For Miniimagenet: We trained our model for this dataset for 15000 iterations. During meta-training and retraining, the inner learning rate is 0.4, and the outer learning rate is 0.001. For meta-testing, the inner learning rate is set to 0.01, and the outer learning rate is set to 0.001. The batch size for both meta-training and meta-testing is set to 16. In both algorithms, only 10 steps of stochastic gradient descent are applied during the evaluation phase. The comparison table shows the classification accuracy (\ref{table:2}).\\

\begin{table}[!h]
\begin{center}
\begin{tabular}{| c | c | c | c | c |}
\hline
\textbf{Method} & \textbf{Prune \%} & \textbf{5way-1shot} &\textbf{5way-5shot}   \\ 
\hline
FOMAML &0 &$48.07\pm1.75$ &$63.15\pm0.91$     \\ \hline
Meta-LTH &90& 45.08 $\pm$ 0.31 & \textbf{64.81 $\pm$ 1.12}   \\ \hline
Meta-LTH &80& 46.00 $\pm$ 1.08 & 58.24 $\pm$ 1.00  \\ \hline
Meta-LTH &70& 46.09 $\pm$ 0.95  & 56.77 $\pm$  1.09  \\ \hline
Meta-LTH &60& 45.94 $\pm$ 1.28 &  56.85 $\pm$ 0.66  \\ \hline
Meta-LTH &50& 46.54 $\pm$ 0.49 &   53.96 $\pm$ 0.17  \\ \hline

\end{tabular}
\caption{ Classification accuracies for proposed Meta-LTH approach in comparison with FOMAML on Miniimagenet dataset.}
\label{table:2}
\end{center}
\end{table}

For Omniglot: We utilized the existing implementation
of FO-MAML in the Torchmeta Python package \cite{deleu2019torchmeta} and adapted it to implement
our Meta-LTH algorithm. During the meta-training phase for the 5-way task setting, we trained the Meta-LTH model until convergence with a learning rate of 0.4 and batch size set to 32. However, during meta-testing, 100 batches of tasks are evaluated with a learning rate of 0.1 and for five gradient steps and 0.01 step size. For the 20-way 1-shot task setting, we trained our proposed model with the same 5-way task training settings, but for 20-way 5-shot tasks, we found that for some pruning percentage settings, we need very less training iterations to get model converge. Rest training and testing settings are similar to the 5-way task settings. The comparison table shows the classification accuracy (\ref{table:3}).\\
\begin{table}[!h]
\begin{center}
\begin{tabular}{| c | c | c | c | c |c|c}
\hline
\textbf{Method} & \textbf{Prune \%} & \textbf{5way-1shot} &\textbf{5way-5shot}& \textbf{20way-1shot}&\textbf{20way-5shot}   \\ 
\hline
FOMAML &0 &$98.3\pm 0.5$ &$99.2\pm0.2$  &$89.4 \pm 0.5$&$97.9 \pm 0.1$\\ \hline
Meta-LTH &90& 97.77 $\pm$ 0.21 & \textbf{99.36 $\pm$ 0.36}   &\textbf{91.68 $\pm$ 0.29} & 97.70 $\pm$ 0.11 \\ \hline
Meta-LTH &80 & 97.74 $\pm$ 0.32 &  \textbf{99.30 $\pm$ 0.06} & \textbf{93.57 $\pm$ 0.17} & \textbf{98.08 $\pm$ 0.04} \\ \hline
Meta-LTH &70&  97.61 $\pm$ 0.81 &  99.29 $\pm$ 0.04  & \textbf{93.75 $\pm$ 0.03} & \textbf{98.36 $\pm$ 0.55}\\ \hline
Meta-LTH &60& 97.52 $\pm$ 0.06 &  99.10 $\pm$ 0.01 & \textbf{93.11 $\pm$ 0.06} & \textbf{98.36 $\pm$ 0.07} \\ \hline
Meta-LTH &50&  96.58 $\pm$ 0.22 &   98.99 $\pm$ 0.21  & \textbf{92.44 $\pm$ 0.01} & \textbf{97.91 $\pm$ 0.05} \\ \hline

\end{tabular}
\caption{ Classification accuracies for proposed Meta-LTH approach in comparison with FOMAML on Omniglot dataset.}
\label{table:3}
\end{center}
\end{table}

In comparison to the benchmark results of the FOMAML algorithm, the proposed Meta-LTH classification result is ideal. Since the benchmark result using the FOMAML algorithm was attained using strict training conditions (60000 iterations), our technique needs extremely low conditions for re-training using the Meta-LTH algorithm, i.e., 6000 iterations for omniglot dataset and CIFAR100 dataset and 15000 iterations for miniimagenet dataset. Because it only uses the necessary weight updation mechanism during the learning phase, the suggested meta-learning approach is more efficient.

\textbf{Meta-learning for Different Test Settings:} We further conduct a set of experiments on Omniglot dataset to better understand the impact of meta-adaptation during meta-testing phase for 5-way 5-shot task setting.

\begin{figure*}[ht]
\centering
\includegraphics[width=9cm,height=6cm]{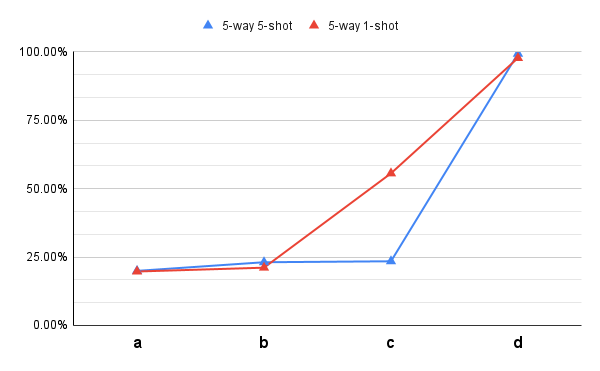}
\caption{\textbf{Graph showing accuracies on omniglot dataset for 5-way task setting:} From left to right, (a) Zero-shot Learning (b)Fine-tuning with only unpruned connections (c) Impact of classifier layer (d) Learning during Meta-LTH }
\label{fig:train}
\end{figure*}
\textbf{(a)Zero-shot Learning:} Once we get the optimal parameters using the sparse model, we need to perform few-shot learning by capturing new features of the unseen task during meta-adaptation. We open some connections in the meta-testing model to learn these new features. These connections are responsible for learning and recombining new unseen task features with the learnt features.
To validate our proposed Meta-LTH method, we performed zero-shot learning on the new unseen tasks. However, the experiments indicate that our Meta-LTH algorithm works far better during the meta-testing phase. Since zero-shot learning experiments produce very subpar few-shot classification findings, the meta-parameters learnt during the meta-training phase do not contribute to identifying the new task.

\textbf{(b)Fine-tuning with only unpruned connections:} Another experimental setting suggests that learning new features with only unpruned connections during the meta-adaptation phase will not lead to a better-generalized model. By restricting the growth of our pruned model during the meta-test phase, we ran experiments to examine how new features are captured during the meta-adaptation phase. The results imply that opening or growing the pruned links in the fine-tuning phase is necessary to capture new unseen features.

\textbf{(c)Impact of classifier layer:} Meta-LTH considers the classifier layer crucial when observed during the meta-training part, as we did not choose this layer to get sparsified during pruning, resulting in rapid convergence during meta-training. However, we only permitted the classifier layer to be updated in anticipation of observing meta-adaptation during meta-testing. The findings indicate that only allowing the classifier layer does not result in feature learning and recombination.

\textbf{(d)Learning during Meta-LTH:} With the success of the proposed method, Meta-LTH observes that during the fine-tuning phase, the initial layers of the model capture the maximum information w.r.t other layers. The weights of the initial convolution layers show low-level learning of new features, whereas the intermediate or last layer weights show high-level learning. Moreover, the new features learnt in the grown edges combine with the learnt information, which performs feature learning and recombination.

\section{Conclusion}
Considering the LTH, pruning a trained neural network effectively retains much of the model's information content and the predictive performance, even for meta-learning approaches. However, to capture the subtleties that need to be learnt in the low data meta-test setting, our proposed method re-activates the pruned connections to do new feature learning and feature reuse with the meta learnt prior. This is empirically shown in the results that reopening pruned parameters during meta-adaptation helps the model learn new features in coherence to the already learnt features and hence gives better classification accuracies on benchmark datasets than FOMAML, a state-of-the-art meta-learning approach. To justify these results, we performed several experiments(zero-shot learning) that show that our method only performs well when connections are opened to learn the low-level features. These low-level features are captured in the initial layers of the network. Therefore, during meta-testing, the initial layers show higher values in the weight matrix than the following layers, which shows that the features required to perform recombination are learned initially in the initial layers of the network. 
Finally, we also conclude that our Meta-LTH method converges fast as we get the optimal parameters for all the datasets with approximately 50\% lesser batch updates than the benchmark.

\bibliographystyle{unsrtnat}
\bibliography{references}  

\begin{thebibliography}{48}
\providecommand{\natexlab}[1]{#1}
\providecommand{\url}[1]{\texttt{#1}}
\expandafter\ifx\csname urlstyle\endcsname\relax
  \providecommand{\doi}[1]{doi: #1}\else
  \providecommand{\doi}{doi: \begingroup \urlstyle{rm}\Url}\fi

\bibitem[Finn et~al.(2017)Finn, Abbeel, and Levine]{finn2017model}
Chelsea Finn, Pieter Abbeel, and Sergey Levine.
\newblock Model-agnostic meta-learning for fast adaptation of deep networks.
\newblock In \emph{International conference on machine learning}, pages
  1126--1135. PMLR, 2017.

\bibitem[Nichol et~al.(2018)Nichol, Achiam, and Schulman]{nichol2018first}
Alex Nichol, Joshua Achiam, and John Schulman.
\newblock On first-order meta-learning algorithms.
\newblock \emph{arXiv preprint arXiv:1803.02999}, 2018.

\bibitem[Raghu et~al.(2019)Raghu, Raghu, Bengio, and Vinyals]{raghu2019rapid}
Aniruddh Raghu, Maithra Raghu, Samy Bengio, and Oriol Vinyals.
\newblock Rapid learning or feature reuse? towards understanding the
  effectiveness of maml.
\newblock \emph{arXiv preprint arXiv:1909.09157}, 2019.

\bibitem[Rajeswaran et~al.(2019)Rajeswaran, Finn, Kakade, and
  Levine]{rajeswaran2019meta}
Aravind Rajeswaran, Chelsea Finn, Sham~M Kakade, and Sergey Levine.
\newblock Meta-learning with implicit gradients.
\newblock \emph{Advances in neural information processing systems}, 32, 2019.

\bibitem[Santoro et~al.(2016)Santoro, Bartunov, Botvinick, Wierstra, and
  Lillicrap]{santoro2016meta}
Adam Santoro, Sergey Bartunov, Matthew Botvinick, Daan Wierstra, and Timothy
  Lillicrap.
\newblock Meta-learning with memory-augmented neural networks.
\newblock In \emph{International conference on machine learning}, pages
  1842--1850. PMLR, 2016.

\bibitem[Tiwari et~al.(2022{\natexlab{a}})Tiwari, Gogoi, Verma, and
  Singh]{9986399}
Sambhavi Tiwari, Manas Gogoi, Shekhar Verma, and Krishna~Pratap Singh.
\newblock Meta-learning with hopfield neural network.
\newblock In \emph{2022 IEEE 9th Uttar Pradesh Section International Conference
  on Electrical, Electronics and Computer Engineering (UPCON)}, pages 1--5,
  2022{\natexlab{a}}.
\newblock \doi{10.1109/UPCON56432.2022.9986399}.

\bibitem[Vinyals et~al.(2016)Vinyals, Blundell, Lillicrap, Wierstra,
  et~al.]{vinyals2016matching}
Oriol Vinyals, Charles Blundell, Timothy Lillicrap, Daan Wierstra, et~al.
\newblock Matching networks for one shot learning.
\newblock \emph{Advances in neural information processing systems}, 29, 2016.

\bibitem[Gogoi et~al.(2022)Gogoi, Tiwari, and Verma]{gogoi2022adaptive}
Manas Gogoi, Sambhavi Tiwari, and Shekhar Verma.
\newblock Adaptive prototypical networks.
\newblock \emph{arXiv preprint arXiv:2211.12479}, 2022.

\bibitem[Guo et~al.(2016)Guo, Yao, and Chen]{guo2016dynamic}
Yiwen Guo, Anbang Yao, and Yurong Chen.
\newblock Dynamic network surgery for efficient dnns.
\newblock \emph{Advances in neural information processing systems}, 29, 2016.

\bibitem[Han et~al.(2015{\natexlab{a}})Han, Mao, and Dally]{han2015deep}
Song Han, Huizi Mao, and William~J Dally.
\newblock Deep compression: Compressing deep neural networks with pruning,
  trained quantization and huffman coding.
\newblock \emph{arXiv preprint arXiv:1510.00149}, 2015{\natexlab{a}}.

\bibitem[Hassibi et~al.(1993)Hassibi, Stork, and Wolff]{hassibi1993optimal}
Babak Hassibi, David~G Stork, and Gregory~J Wolff.
\newblock Optimal brain surgeon and general network pruning.
\newblock In \emph{IEEE international conference on neural networks}, pages
  293--299. IEEE, 1993.

\bibitem[LeCun et~al.(1989)LeCun, Denker, and Solla]{lecun1989optimal}
Yann LeCun, John Denker, and Sara Solla.
\newblock Optimal brain damage.
\newblock \emph{Advances in neural information processing systems}, 2, 1989.

\bibitem[Ravi and Larochelle(2017)]{ravi2017optimization}
S~Ravi and H~Larochelle.
\newblock Optimization as a model for few-shot learning. 5th int.
\newblock In \emph{Conf. Learn. Represent. ICLR 2017-Conf. Track Proc. 1--11},
  2017.

\bibitem[Elesedy et~al.(2020)Elesedy, Kanade, and Teh]{elesedy2020lottery}
Bryn Elesedy, Varun Kanade, and Yee~Whye Teh.
\newblock Lottery tickets in linear models: An analysis of iterative magnitude
  pruning.
\newblock \emph{arXiv preprint arXiv:2007.08243}, 2020.

\bibitem[Oh et~al.(2020)Oh, Yoo, Kim, and Yun]{oh2020boil}
Jaehoon Oh, Hyungjun Yoo, ChangHwan Kim, and Se-Young Yun.
\newblock Boil: Towards representation change for few-shot learning.
\newblock \emph{arXiv preprint arXiv:2008.08882}, 2020.

\bibitem[Koch et~al.(2015)Koch, Zemel, Salakhutdinov, et~al.]{koch2015siamese}
Gregory Koch, Richard Zemel, Ruslan Salakhutdinov, et~al.
\newblock Siamese neural networks for one-shot image recognition.
\newblock In \emph{ICML deep learning workshop}, volume~2. Lille, 2015.

\bibitem[Snell et~al.(2017)Snell, Swersky, and Zemel]{snell2017prototypical}
Jake Snell, Kevin Swersky, and Richard Zemel.
\newblock Prototypical networks for few-shot learning.
\newblock \emph{Advances in neural information processing systems}, 30, 2017.

\bibitem[Sung et~al.(2018)Sung, Yang, Zhang, Xiang, Torr, and
  Hospedales]{sung2018learning}
Flood Sung, Yongxin Yang, Li~Zhang, Tao Xiang, Philip~HS Torr, and Timothy~M
  Hospedales.
\newblock Learning to compare: Relation network for few-shot learning.
\newblock In \emph{Proceedings of the IEEE conference on computer vision and
  pattern recognition}, pages 1199--1208, 2018.

\bibitem[Mishra et~al.(2017)Mishra, Rohaninejad, Chen, and
  Abbeel]{mishra2017simple}
Nikhil Mishra, Mostafa Rohaninejad, Xi~Chen, and Pieter Abbeel.
\newblock A simple neural attentive meta-learner.
\newblock \emph{arXiv preprint arXiv:1707.03141}, 2017.

\bibitem[ur~Rehman et~al.(2023)ur~Rehman, Ali, Jan, Ali, Xu, and
  Shao]{ur2023caml}
Israr ur~Rehman, Waqar Ali, Zahoor Jan, Zulfiqar Ali, Hui Xu, and Jie Shao.
\newblock Caml: Contextual augmented meta-learning for cold-start
  recommendation.
\newblock \emph{Neurocomputing}, 533:\penalty0 178--190, 2023.

\bibitem[Antoniou et~al.(2018)Antoniou, Edwards, and
  Storkey]{antoniou2018train}
Antreas Antoniou, Harrison Edwards, and Amos Storkey.
\newblock How to train your maml.
\newblock \emph{arXiv preprint arXiv:1810.09502}, 2018.

\bibitem[Nichol and Schulman(2018)]{nichol2018reptile}
Alex Nichol and John Schulman.
\newblock Reptile: a scalable metalearning algorithm.
\newblock \emph{arXiv preprint arXiv:1803.02999}, 2\penalty0 (3):\penalty0 4,
  2018.

\bibitem[Elsken et~al.(2020)Elsken, Staffler, Metzen, and
  Hutter]{elsken2020meta}
Thomas Elsken, Benedikt Staffler, Jan~Hendrik Metzen, and Frank Hutter.
\newblock Meta-learning of neural architectures for few-shot learning.
\newblock In \emph{Proceedings of the IEEE/CVF conference on computer vision
  and pattern recognition}, pages 12365--12375, 2020.

\bibitem[Hou and Kwok(2018)]{hou2018loss}
Lu~Hou and James~T Kwok.
\newblock Loss-aware weight quantization of deep networks.
\newblock \emph{arXiv preprint arXiv:1802.08635}, 2018.

\bibitem[Zhuang et~al.(2019)Zhuang, Shen, Tan, Liu, and
  Reid]{zhuang2019structured}
Bohan Zhuang, Chunhua Shen, Mingkui Tan, Lingqiao Liu, and Ian Reid.
\newblock Structured binary neural networks for accurate image classification
  and semantic segmentation.
\newblock In \emph{Proceedings of the IEEE/CVF Conference on Computer Vision
  and Pattern Recognition}, pages 413--422, 2019.

\bibitem[Zhou et~al.(2018)Zhou, Yao, Wang, and Chen]{zhou2018explicit}
Aojun Zhou, Anbang Yao, Kuan Wang, and Yurong Chen.
\newblock Explicit loss-error-aware quantization for low-bit deep neural
  networks.
\newblock In \emph{Proceedings of the IEEE conference on computer vision and
  pattern recognition}, pages 9426--9435, 2018.

\bibitem[Sandler et~al.(2018)Sandler, Howard, Zhu, Zhmoginov, and
  Chen]{sandler2018mobilenetv2}
Mark Sandler, Andrew Howard, Menglong Zhu, Andrey Zhmoginov, and Liang-Chieh
  Chen.
\newblock Mobilenetv2: Inverted residuals and linear bottlenecks.
\newblock In \emph{Proceedings of the IEEE conference on computer vision and
  pattern recognition}, pages 4510--4520, 2018.

\bibitem[Zhang et~al.(2018)Zhang, Zhou, Lin, and Sun]{zhang2018shufflenet}
Xiangyu Zhang, Xinyu Zhou, Mengxiao Lin, and Jian Sun.
\newblock Shufflenet: An extremely efficient convolutional neural network for
  mobile devices.
\newblock In \emph{Proceedings of the IEEE conference on computer vision and
  pattern recognition}, pages 6848--6856, 2018.

\bibitem[Iandola et~al.(2016)Iandola, Han, Moskewicz, Ashraf, Dally, and
  Keutzer]{iandola2016squeezenet}
Forrest~N Iandola, Song Han, Matthew~W Moskewicz, Khalid Ashraf, William~J
  Dally, and Kurt Keutzer.
\newblock Squeezenet: Alexnet-level accuracy with 50x fewer parameters and< 0.5
  mb model size.
\newblock \emph{arXiv preprint arXiv:1602.07360}, 2016.

\bibitem[He et~al.(2017)He, Zhang, and Sun]{he2017channel}
Yihui He, Xiangyu Zhang, and Jian Sun.
\newblock Channel pruning for accelerating very deep neural networks.
\newblock In \emph{Proceedings of the IEEE international conference on computer
  vision}, pages 1389--1397, 2017.

\bibitem[Liu et~al.(2019)Liu, Mu, Zhang, Guo, Yang, Cheng, and
  Sun]{liu2019metapruning}
Zechun Liu, Haoyuan Mu, Xiangyu Zhang, Zichao Guo, Xin Yang, Kwang-Ting Cheng,
  and Jian Sun.
\newblock Metapruning: Meta learning for automatic neural network channel
  pruning.
\newblock In \emph{Proceedings of the IEEE/CVF international conference on
  computer vision}, pages 3296--3305, 2019.

\bibitem[Tian et~al.(2020)Tian, Liu, Yuan, and Liu]{tian2020meta}
Hongduan Tian, Bo~Liu, Xiao-Tong Yuan, and Qingshan Liu.
\newblock Meta-learning with network pruning.
\newblock In \emph{European Conference on Computer Vision}, pages 675--700.
  Springer, 2020.

\bibitem[Han et~al.(2016)Han, Pool, Narang, Mao, Gong, Tang, Elsen, Vajda,
  Paluri, Tran, et~al.]{han2016dsd}
Song Han, Jeff Pool, Sharan Narang, Huizi Mao, Enhao Gong, Shijian Tang, Erich
  Elsen, Peter Vajda, Manohar Paluri, John Tran, et~al.
\newblock Dsd: Dense-sparse-dense training for deep neural networks.
\newblock \emph{arXiv preprint arXiv:1607.04381}, 2016.

\bibitem[Jin et~al.(2016)Jin, Yuan, Feng, and Yan]{jin2016training}
Xiaojie Jin, Xiaotong Yuan, Jiashi Feng, and Shuicheng Yan.
\newblock Training skinny deep neural networks with iterative hard thresholding
  methods.
\newblock \emph{arXiv preprint arXiv:1607.05423}, 2016.

\bibitem[Munkhdalai and Yu(2017)]{munkhdalai2017meta}
Tsendsuren Munkhdalai and Hong Yu.
\newblock Meta networks.
\newblock In \emph{International conference on machine learning}, pages
  2554--2563. PMLR, 2017.

\bibitem[Tiwari et~al.(2022{\natexlab{b}})Tiwari, Gogoi, Verma, and
  Singh]{tiwari2022meta}
Sambhavi Tiwari, Manas Gogoi, Shekhar Verma, and Krishna~Pratap Singh.
\newblock Meta-learning with hopfield neural network.
\newblock In \emph{2022 IEEE 9th Uttar Pradesh Section International Conference
  on Electrical, Electronics and Computer Engineering (UPCON)}, pages 1--5.
  IEEE, 2022{\natexlab{b}}.

\bibitem[Frankle and Carbin(2018)]{frankle2018lottery}
Jonathan Frankle and Michael Carbin.
\newblock The lottery ticket hypothesis: Finding sparse, trainable neural
  networks.
\newblock \emph{arXiv preprint arXiv:1803.03635}, 2018.

\bibitem[Wen et~al.(2016)Wen, Wu, Wang, Chen, and Li]{wen2016learning}
Wei Wen, Chunpeng Wu, Yandan Wang, Yiran Chen, and Hai Li.
\newblock Learning structured sparsity in deep neural networks.
\newblock \emph{Advances in neural information processing systems}, 29, 2016.

\bibitem[Han et~al.(2015{\natexlab{b}})Han, Pool, Tran, and
  Dally]{han2015learning}
Song Han, Jeff Pool, John Tran, and William Dally.
\newblock Learning both weights and connections for efficient neural network.
\newblock \emph{Advances in neural information processing systems}, 28,
  2015{\natexlab{b}}.

\bibitem[Li et~al.(2016)Li, Kadav, Durdanovic, Samet, and Graf]{li2016pruning}
Hao Li, Asim Kadav, Igor Durdanovic, Hanan Samet, and Hans~Peter Graf.
\newblock Pruning filters for efficient convnets.
\newblock \emph{arXiv preprint arXiv:1608.08710}, 2016.

\bibitem[Liu et~al.(2018)Liu, Sun, Zhou, Huang, and Darrell]{liu2018rethinking}
Zhuang Liu, Mingjie Sun, Tinghui Zhou, Gao Huang, and Trevor Darrell.
\newblock Rethinking the value of network pruning.
\newblock \emph{arXiv preprint arXiv:1810.05270}, 2018.

\bibitem[Bai et~al.(2022)Bai, Wang, Tao, Li, and Fu]{bai2022dual}
Yue Bai, Huan Wang, Zhiqiang Tao, Kunpeng Li, and Yun Fu.
\newblock Dual lottery ticket hypothesis.
\newblock \emph{arXiv preprint arXiv:2203.04248}, 2022.

\bibitem[Burkholz et~al.(2021)Burkholz, Laha, Mukherjee, and
  Gotovos]{burkholz2021existence}
Rebekka Burkholz, Nilanjana Laha, Rajarshi Mukherjee, and Alkis Gotovos.
\newblock On the existence of universal lottery tickets.
\newblock \emph{arXiv preprint arXiv:2111.11146}, 2021.

\bibitem[Dong et~al.(2017)Dong, Chen, and Pan]{dong2017learning}
Xin Dong, Shangyu Chen, and Sinno Pan.
\newblock Learning to prune deep neural networks via layer-wise optimal brain
  surgeon.
\newblock \emph{Advances in Neural Information Processing Systems}, 30, 2017.

\bibitem[Lake et~al.(2015)Lake, Salakhutdinov, and Tenenbaum]{lake2015human}
Brenden~M Lake, Ruslan Salakhutdinov, and Joshua~B Tenenbaum.
\newblock Human-level concept learning through probabilistic program induction.
\newblock \emph{Science}, 350\penalty0 (6266):\penalty0 1332--1338, 2015.

\bibitem[Oreshkin et~al.(2018)Oreshkin, Rodr{\'\i}guez~L{\'o}pez, and
  Lacoste]{oreshkin2018tadam}
Boris Oreshkin, Pau Rodr{\'\i}guez~L{\'o}pez, and Alexandre Lacoste.
\newblock Tadam: Task dependent adaptive metric for improved few-shot learning.
\newblock \emph{Advances in neural information processing systems}, 31, 2018.

\bibitem[Deleu et~al.(2019)Deleu, W{\"u}rfl, Samiei, Cohen, and
  Bengio]{deleu2019torchmeta}
Tristan Deleu, Tobias W{\"u}rfl, Mandana Samiei, Joseph~Paul Cohen, and Yoshua
  Bengio.
\newblock Torchmeta: A meta-learning library for pytorch.
\newblock \emph{arXiv preprint arXiv:1909.06576}, 2019.

\bibitem[Krizhevsky et~al.(2009)Krizhevsky, Hinton,
  et~al.]{krizhevsky2009learning}
Alex Krizhevsky, Geoffrey Hinton, et~al.
\newblock Learning multiple layers of features from tiny images.
\newblock 2009.

\end{thebibliography}






\end{document}